\documentclass[preprint,12pt]{elsarticle}
\usepackage{booktabs}
\usepackage{subcaption}
\usepackage[colorlinks=true, linkcolor=blue, citecolor=blue, urlcolor=blue]{hyperref}
\hypersetup{colorlinks=true}

\usepackage{amssymb}
\usepackage{amsmath}

\journal{Magnetic Resonance Imaging}

\begin{document}

\begin{frontmatter}
	
\title{EffNetViTLoRA: An Efficient Hybrid Deep Learning Approach for Alzheimer’s Disease Diagnosis}
\author[1]{Mahdieh Behjat Khatooni}
\ead{mahdiyeh.bhjt@gmail.com}

\author[2]{Mohsen Soryani\corref{cor1}}
\ead{soryani@iust.ac.ir}

\cortext[cor1]{Corresponding author}

\affiliation[1]{
	organization={School of Computer Engineering, Iran University of Science and Technology},
	city={Tehran},
	state={Tehran},
	country={Iran}
}

\affiliation[2]{
	organization={School of Computer Engineering, Iran University of Science and Technology},
	city={Tehran},
	state={Tehran},
	country={Iran}
}
\begin{abstract}
Alzheimer’s disease (AD) is one of the most prevalent neurodegenerative disorders worldwide. As it progresses, it leads to the deterioration of cognitive functions. Since AD is irreversible, early diagnosis is crucial for managing its progression. Mild Cognitive Impairment (MCI) represents an intermediate stage between Cognitively Normal (CN) individuals and those with AD, and is considered a transitional phase from normal cognition to Alzheimer’s disease. Diagnosing MCI is particularly challenging due to the subtle differences between adjacent diagnostic categories. In this study, we propose EffNetViTLoRA, a generalized end-to-end model for AD diagnosis using the whole Alzheimer’s Disease Neuroimaging Initiative (ADNI) Magnetic Resonance Imaging (MRI) dataset. Our model integrates a Convolutional Neural Network (CNN) with a Vision Transformer (ViT) to capture both local and global features from MRI images. Unlike previous studies that rely on limited subsets of data, our approach is trained on the full T1-weighted MRI dataset from ADNI, resulting in a more robust and unbiased model. This comprehensive methodology enhances the model’s clinical reliability. Furthermore, fine-tuning large pretrained models often yields suboptimal results when source and target dataset domains differ. To address this, we incorporate Low-Rank Adaptation (LoRA) to effectively adapt the pretrained ViT model to our target domain. This method enables efficient knowledge transfer and reduces the risk of overfitting. Our model achieves a classification accuracy of 92.52\% and an F1-score of 92.76\% across three diagnostic categories: AD, MCI, and CN for full ADNI dataset.
\end{abstract}

\begin{highlights}
\item Proposed diagnosis model classifies Alzheimer’s disease into AD, MCI, and CN
\item Proposed hybrid diagnosis model combines a CNN and a ViT to extract MRI features
\item Our model uses ALL ADNI T1-weighted MRI data, unlike previous studies using subsets
\item Using unbiased data, our model offers more reliable AI-assisted diagnosis
\item Using LoRA, our model achieves 92.52\% accuracy with only thousands of parameters
\end{highlights}

\begin{keyword}
	Alzheimer’s disease \sep 
	deep learning \sep
	diagnosis \sep
	LoRA \sep
	MRI \sep
	Vision Transformer

\end{keyword}

\end{frontmatter}



\section{Introduction}
\label{sec:introduction}
Alzheimer's disease (AD) is the most common irreversible neurodegenerative disorder leading to dementia worldwide. It is characterized by the destruction of neurons in specific regions of the brain specifically hippocampi, resulting in significant disruptions to cognitive and perceptual functions. As AD progresses, cognitive decline intensifies, initially manifesting as forgetfulness and memory loss. Over time, individuals lose the ability to perform even simple tasks, eventually becoming entirely dependent on others for basic daily activities \cite{AlzheimerSymptoms}. The disease ultimately results in death. Given its irreversible nature, early diagnosis of AD is increasingly critical for effective management and intervention. The number of elderly individuals affected by this disease is growing significantly each year. The community is likely to rise to about 152 million people by 2050. The current annual cost of managing AD is about one trillion USD and is expected to double by 2030 \cite{patterson2018world}. Despite extensive research, the exact cause of AD remains unknown and there is not a definitive cure for that.

Recently, Convolutional Neural Networks (CNNs) and Vision Transformers (ViTs) \cite{dosovitskiy2020image} have gained prominence in medical image processing and analysis. These advancements have made substantial contributions to diagnostic efforts by enabling the automated analysis of structural and functional patterns in medical images, as well as non-imaging clinical data. Among various imaging modalities, Magnetic Resonance Imaging (MRI) is particularly favored due to its non-invasive nature and ability to capture high-resolution structural details of the brain.

In this paper, we review recent research on Alzheimer's disease diagnosis, covering both binary and multi-class classification approaches, in Section~\ref{sec:related-works}. Section~\ref{sec:method} provides a detailed description of the proposed hybrid diagnostic architecture. In Section~\ref{sec:experiments}, we introduce the dataset and preprocessing steps, followed by the presentation and analysis of our model's classification results. Finally, Section~\ref{sec:conclusion} summarizes the key findings and concludes the study.

\section{Related Work}
\label{sec:related-works}
Alzheimer’s Disease is typically categorized into three levels by clinicians and researchers based on progression: 1. Cognitively Normal (CN): Individuals with no signs of cognitive impairment. 2. Mild Cognitive Impairment (MCI): An intermediate stage between CN and AD that is often targeted for early intervention to prevent progression to AD. 3. AD: Patients who exhibit full clinical symptoms of Alzheimer’s disease. Many of the studies published on Alzheimer’s disease diagnosis have focused on binary classification, while only a few have proposed multi-class classification approaches that include MCI.

\subsection{Works on classification of CN vs AD}
In \cite{jang2022m3t}, the hybrid M3T model, combines a 3D CNN, a multi-plane 2D feature extractor with a pretrained ResNet50 \cite{he2016deep} and a Vision Transformer. It could achieve an accuracy of 93.21\% on ADNI. In \cite{li2022trans}, the Trans-ResNet model combines a 3D CNN and a ViT with transfer learning from brain age estimation to achieve over 93\% accuracy. In \cite{liu2023cascaded}, the 3MT model integrates 13 multimodal data types using cascaded attention \cite{vaswani2017attention} mechanisms, achieving 99\% accuracy through a unified representation framework. In \cite{qiu2020development}, a deep learning framework combines a 3D fully convolutional network (FCN) and multi layer perceptron (MLP) to classify Alzheimer's disease by extracting high-correlation voxel features from MRI data, validated across multiple datasets. In \cite{ge2019multi}, a multi-stream, multi-scale 3D CNN extracts region-specific brain features, enhanced by cross-region fusion, XGBoost \cite{chen2016xgboost} for dimensionality reduction, and transfer learning for improved accuracy. In \cite{liu2020multi}, a two-stage model integrates hippocampus segmentation with DenseNet \cite{huang2017densely} to refine feature extraction to enhance classification through fused multi-stage features. In \cite{zhang2022diagnosis}, a ResNet model uses segmented gray matter regions from 2D MRIs, enhanced by space to depth resolution reduction and channel attention.

\subsection{Works on multi class classification of AD}
In \cite{ding2019deep}, a fine-tuned Inception V3 \cite{szegedy2016rethinking} model analyzes preprocessed F-FDG PET scans for AD diagnosis, outperforming clinical experts. \cite{liu2022diagnosis} employs a modified ResNet-50 with multi-scale residual blocks and Squeeze-and-Excitation attention \cite{hu2018squeeze} to diagnose AD using segmented white matter and gray matter MRI slices. In \cite{qin20223d}, a UNet-inspired model \cite{ronneberger2015u} processes 3D MRI images using a hybrid attention mechanism with downsampling-upsampling, combining 3D channel and spatial attention to preserve spatial details and highlight relevant features for AD diagnosis. In \cite{zhang20213d}, a 3D DenseNet-based framework \cite{zhang20213d} is proposed using 3D MRI images, replacing 2D convolutions with 3D ones. An attention mechanism is incorporated, dynamically weighting features based on their relevance. \cite{zhang2021explainable} combines voxel-level and region-specific features hierarchically. Interpretability is achieved using Grad-CAM \cite{selvaraju2017grad}. \cite{zhu2021dual} proposes a dual-attention deep learning model that combines multi-instance learning with spatial attention and impact coefficients to localize pathological regions from MRI patches for accurate AD diagnosis. In \cite{lin2021multiclass}, the authors proposed a multimodal framework combining LDA (Linear Discriminant Analysis) for feature extraction with an extreme learning machine-based decision tree to classify AD stages, achieving 66.7\% accuracy in a three-class setting. In \cite{altaf2018multi}, a machine learning framework is proposed, combining features from MRI with clinical biomarkers. Feature selection techniques are utilized, and multiple classifiers are applied. \cite{khan2019transfer} employs a pre-trained VGG-19 \cite{simonyan2014very} and a novel training data selection strategy to enhance Alzheimer's disease classification accuracy using MRI data. \cite{golovanevsky2022multimodal} introduces the Multimodal Alzheimer's Disease Diagnosis framework, which integrates imaging, genetic, and clinical data, and employs cross-modal attention mechanisms to capture interactions between diverse data types. In \cite{helaly2022deep}, classification is performed with 2D and 3D convolutional networks. The architectures are similar, differing only in kernel and pooling dimensions. 

\section{Method}
\label{sec:method}
\subsection{EffNetViTLoRA}
\begin{figure}
	\centering
	\includegraphics[width=\columnwidth]{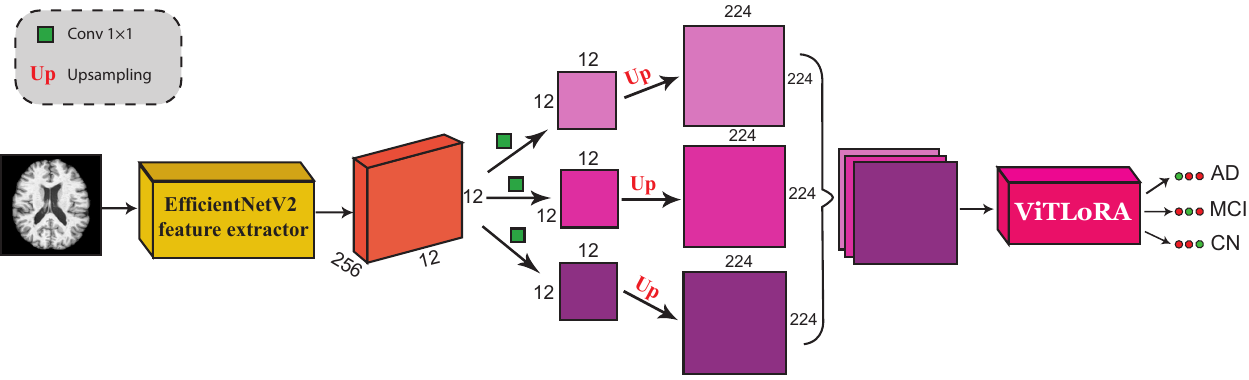}
	\caption{Architecture of EffNetViTLoRA. The hybrid model captures both local and global dependencies in input MRI images by combining Convolutional Neural Networks with a Vision Transformer.}
	\label{fig:hybrid}
\end{figure}

In this section, we introduce our hybrid model, EffNetViTLoRA, which integrates an EfficientNetV2 small \cite{tan2021efficientnetv2} with ViTLoRA. ViTLoRA adapts a pretrained Vision Transformer using Low-Rank Adaptation (LoRA) \cite{hu2021lora} matrices. Fig.~\ref{fig:hybrid} provides an abstract illustration of EffNetViTLoRA.
To leverage both local and global representations from MRI images, we proposed a hybrid model combining EfficientNetV2 Small and ViTLoRA which are described in subsequent sections. We used EfficientNetV2 Small, pretrained on the ImageNet-1K \cite{deng2009imagenet} dataset, as a local feature extractor to generate representative feature maps. Specifically, we extract the feature maps from the output of the 15th MBConv (Mobile Inverted Bottleneck Convolution) \cite{sandler2018mobilenetv2} block at the final stage of EfficientNetV2, which consists of 256 channels. To adapt these feature maps as input to the ViTLoRA basic pretrained model, we applied three 1 × 1 convolutional kernels to reduce the number of channels to 3. Next, to ensure compatibility with the input dimensions of ViTLoRA, we used upsampling to scale the feature maps to 224 × 224. These preprocessed feature maps are then fed into ViTLoRA, to capture global features and dependencies. By combining EfficientNet for local feature extraction with ViTLoRA for global attention-based representations, our hybrid end-to-end diagnostic model leverages the strengths of both architectures.
\subsection{EfficientNet}
EfficientNet, introduced by Tan and Le, focuses on scaling CNN architectures efficiently by balancing depth, width, and resolution. EfficientNetV2 builds on its predecessor by introducing further optimizations to reduce FLOPs (Floating Point Operations Per Second), trainable parameters, and training time. A significant change in EfficientNetV2 is the replacement of MBConv layers in earlier stages with Fused-MBConv layers \cite{gupta2019efficientnet}. Unlike MBConv, which uses a 1 × 1 convolution followed by a depthwise 3 × 3 convolution, Fused-MBConv simplifies this structure by employing a regular 3 × 3 convolution. Depthwise convolutions have fewer parameters and FLOPs than regular convolutions, but they often cannot fully utilize modern accelerators \cite{tan2021efficientnetv2}. This modification addresses a limitation of depthwise convolutions. While they reduce parameters and FLOPs, they often fail to fully utilize the computational potential of modern hardware accelerators. By incorporating Fused-MBConv layers, EfficientNetV2 achieves a better balance between computational efficiency and performance, making it a state-of-the-art architecture for tasks requiring effective feature extraction with optimized training resources. We selected this architecture due to its innovative design, which balances performance and computational efficiency, and its proven effectiveness on the ImageNet benchmark.
\subsection{Vision Transformer (ViT)}
\begin{figure}
	\centering
	\includegraphics[width=\textwidth]{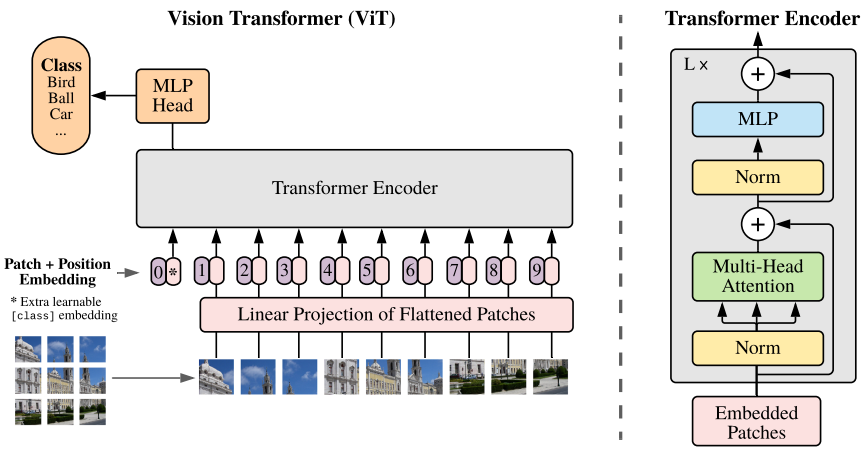}
	\caption{Original architecture of the ViT \cite{dosovitskiy2020image}.}
	\label{fig:ViT}
\end{figure}

Transformers were initially introduced in the field of NLP (Natural Language Processing), where they process tokenized input sequences through layers of encoders and decoders to predict the next token. In 2020, the Vision Transformer was proposed for image-related tasks, adapting the transformer architecture for computer vision applications. The architecture of ViT is shown in Fig.~\ref{fig:ViT}.
The ViT architecture leverages the self-attention mechanism, inspired by the human visual system \cite{liu2024vision}, to understand the relationships between image patches. In ViT, an input image is divided into small, non-overlapping patches. These patches are then flattened to create a sequence of tokens. After positional encodings are added to each token to preserve their order, the tokens are passed through multiple encoder blocks. Each encoder contains a multi-head self-attention mechanism, which computes different attention-based features by modeling the relationships between all tokens in the sequence. Equation~\ref{formul_attention} presents the calculation of attention \cite{vaswani2017attention}, where Q, K and V stand for the query, key, and value weight matrices, respectively, and $d_k$ is the dimension of the key vector used for normalization.

This approach enables the extraction of global features, as it considers the entire image context during processing. Such global attention allows ViT models to effectively capture dependencies across the entire image, making them particularly well-suited for tasks requiring high-level contextual understanding, such as classification and object detection.
\begin{equation}
	\label{formul_attention}
	\text{Attention}(Q, K, V) = \text{softmax}\left(\frac{QK^\top}{\sqrt{d_k}}\right)V  
\end{equation}
\subsection{Low Rank Adaptation (LoRA)}
Training large, high-performance deep models often requires substantial computational resources and training time. Another challenge, particularly in medical applications, is the limited availability of data, which can hinder the development of reliable models.	
One widely used approach to address the issue of limited data is transfer learning. This technique initializes the model's weights based on a pretrained task, providing a good starting point for training on small datasets, such as those in medical applications. Transfer learning is especially effective for adapting large deep learning models to medical datasets. However, fine-tuning large models with millions of parameters is time-intensive and computationally expensive. Additionally, modifying millions of parameters simultaneously may not guarantee optimal performance on the new task. An alternative method to adapt pretrained models to target datasets is Low-Rank Adaptation. LoRA introduces low-rank decomposition matrices into the model, allowing efficient parameter updates while keeping the original pretrained weights largely intact. Indeed, LoRA attempts to learn the differences between the source dataset and the target dataset. It learns small low-rank matrices in certain layers of the model, keeping the pretrained weights frozen. This method significantly reduces the number of trainable parameters, making the adaptation process faster and more resource-efficient. By leveraging LoRA, models can be fine-tuned with limited computational resources, making it an attractive approach for scenarios where time and data availability are constrained.
\subsection{Vision Transformer Low Rank Adaptation (ViTLoRA)}
\begin{figure*}
	\centering
	\includegraphics[width=\textwidth]{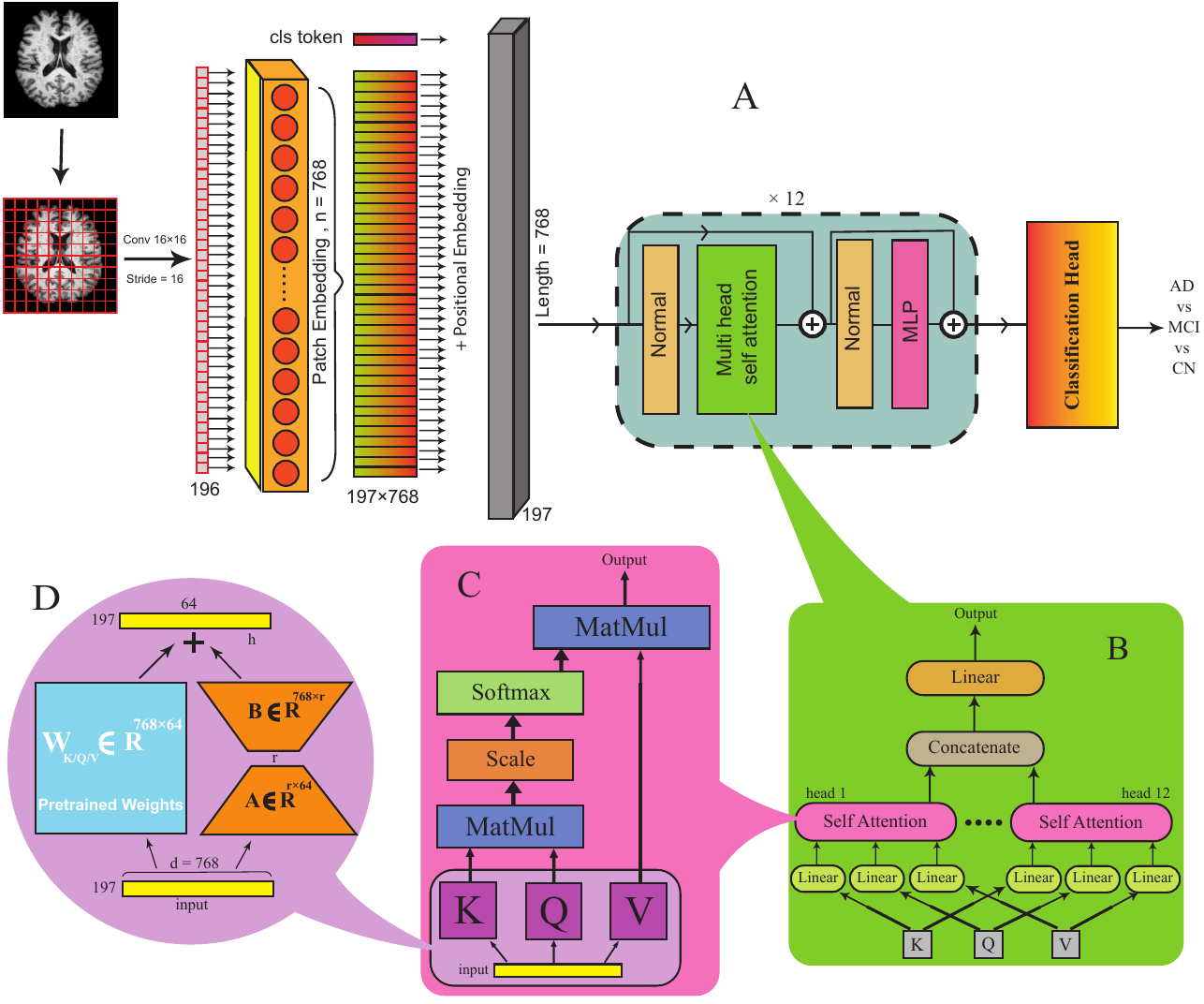}
	\caption{The architecture of ViTLoRA. In each encoder block, we applied low-rank decomposable matrices A and B to the key, query, and value weight matrices to learn the differences between the source and target datasets. Part A illustrates the overall architecture flow. Part B shows the structure of a Multi-Head Self-Attention module. Part C details the architecture of a self-attention block. Part D highlights where the low-rank adaptation matrices A and B are injected into the K, Q and V weight matrices within each self-attention module.}
	\label{fig:ViTLoRA}
\end{figure*}

Training Vision Transformers on medical datasets poses a significant challenge due to the limited availability of labeled data. ViTs, as large-scale models, require substantial data to perform effectively. To address this limitation, we propose incorporating LoRA matrices into the base ViT model (ViT-B/16, using 16×16 patches) to adapt the model for Alzheimer's disease classification using MRI data. Fig.~\ref{fig:ViTLoRA} provides a detailed illustration of the ViTLoRA architecture used in our proposed method. Our approach leverages the differences between the ImageNet-1K dataset and the ADNI dataset to enhance the model's adaptability.
We utilized LoRA matrices with a rank of 4 to adapt the key, query, and value matrices in each encoder of the ViT. Additionally, we added a custom classification head with one hidden layer comprising 256 units, followed by a SoftMax layer with three output neurons to accommodate classification into three classes (AD vs. MCI vs. CN). We refer to this modified model as ViTLoRA.
Since we apply LoRA to a pretrained base ViT, the input to the self-attention layer (197 × 768) follows two branches, as shown in part D of Fig.~\ref{fig:ViTLoRA}. In the first branch, it is multiplied by the pretrained Key, Query, and Value weight matrices (768 × 64). In the second branch, two low-rank decomposition matrices, A and B, are introduced, with sizes 4 × 64 and 768 × 4, respectively. These matrices are designed to capture the differences between the source and target datasets, while keeping the pretrained Key, Query, and Value weights frozen. Equation \ref{formul_LoRA} \cite{hu2021lora} represents the computation of LoRA. The product of A and B yields an abstract difference matrix, $\Delta W$, which is compatible with the output of the first branch. $\Delta W$ is then multiplied by the same input matrix in the right branch. The outputs from both branches are summed to form the updated Key, Query, and Value matrices ($h$). There are three distinct $\Delta W$ matrices corresponding to the three learned weight matrices in each self-attention block, namely: $\Delta W_K$, $\Delta W_Q$, and $\Delta W_V$.	

\begin{equation}
	\label{formul_LoRA}
	h = Wx + \Delta Wx = Wx + BAx
\end{equation}

The output CLS token from the 12th encoder block, which contains a summary of the image patch characteristics, is fed into the classification head. This head comprises one hidden layer with 256 neurons using ReLU activation, followed by an output layer with 3 neurons using a softmax activation function for three-class classification.

\section{Experiments}
\label{sec:experiments}
\subsection{Dataset}
\begin{figure}
	\centering
	\includegraphics[scale=1,width=\columnwidth]{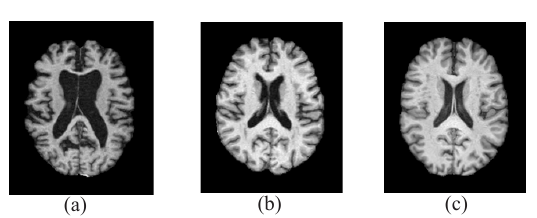}
	\caption{A sample slice of axial plane from ADNI for each class of Alzheimer's disease diagnosis. (a) AD. (b) MCI. (c) CN.}
	\label{fig:3classes}
\end{figure}

In this work, we used data from the Alzheimer’s Disease Neuroimaging Initiative (ADNI)\footnote{\href{http://adni.loni.usc.edu}{http://adni.loni.usc.edu}}. ADNI was launched in 2003 as a public-private partnership, led by Principal Investigator Michael W. Weiner, MD. The primary goal of ADNI has been to test whether serial MRI, positron emission tomography (PET), other biological markers, and clinical and neuropsychological assessments can be combined to measure the progression of mild cognitive impairment and early Alzheimer’s disease. We utilized 2010 subjects, all T1-weighted MRI volumes available in the ADNI1, ADNIGO, ADNI2, and ADNI3 phases at the time of our experiments. Table \ref{tab:demographics} presents the detailed demographics and clinical information of the included subjects. While multiple MRI scans are available for most subjects, representing time series data, we used only the baseline visit volumes of each subject for consistency.
Sample images from each of three classes are illustrated in Fig.~\ref{fig:3classes}. Although we tried to select and display images with obvious differences in each of the categories in Fig.~\ref{fig:3classes}, the difficulty of distinguishing between MCI and CN cases is evident, which presents a significant challenge for diagnosis and model training in three classes. Atrophy in the GM and hippocampal regions of the medial temporal lobe are among the most noticeable effects of Alzheimer’s disease \cite{cavedo2014medial}.
\begin{table}
	\caption{Summary of the demographic information of the included subjects across all ADNI phases.}
	\centering
	\label{tab:demographics}
	\begin{tabular}{@{}llll@{}}
		\toprule
		\textbf{Category}       & 
		\textbf{Number of subjects}    &
		\textbf{Gender(M/F)}    &
		\textbf{Age(years) Mean ± SD} \\ \midrule
		AD             & 444      & 244/200         & 75.88 ± 07.91    \\
		MCI                  & 706      & 415/291       & 76.07 ± 07.68        \\
		CN        & 860     & 375/485          & 76.28 ± 06.79     \\
		Combined         &    2010    & 1034/976       & 76.11 ± 07.36  
		\\ \bottomrule
	\end{tabular}
\end{table}

\begin{figure}
	\centering
	\includegraphics[width=\columnwidth]{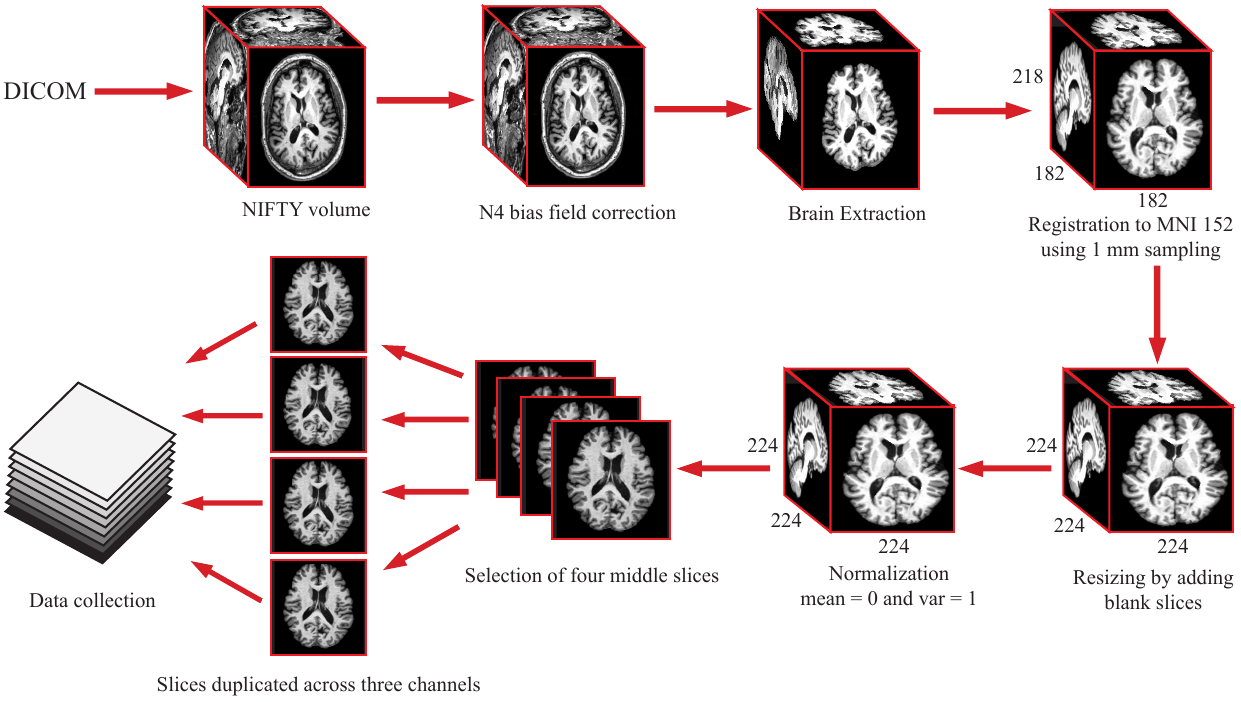}
	\caption{Preprocessing steps of ADNI MRI T1-weighted volumes in our study. The arrows demonstrate the flow of the process.}
	\label{fig:preprocess}
\end{figure}

Preprocessing steps is shown in Fig.~\ref{fig:preprocess}. DICOM layers are converted to NIfTI format using MRICron software. To ensure consistent image quality, we applied N4 bias field correction \cite{tustison2010n4itk} to standardize image intensity. Next, Brain Extraction Tool (BET) \cite{isensee2019automated} was used to remove non-brain tissue, isolating brain anatomy and enabling the model to focus exclusively on relevant structures. Each volume was spatially normalized to the MNI152 template space with a 1mm sampling resolution using the FLIRT \cite{jenkinson2001global,jenkinson2002improved} registration tool in the FSL \cite{woolrich2009bayesian} software package, resulting in volumes of size 218 × 182 × 218. To standardize dimensions for compatibility with the input requirements of pretrained Vision Transformer models, we added blank slices to the beginning and end of each volume, adjusting the dimensions to 224 × 224 × 224. Finally, each volume was normalized to have a zero mean and unit variance, ensuring consistent voxel value ranges. Preprocessing includes selecting representative 2D slices from each volume. From the center of each volume along the axial plane, we extracted four consecutive middle slices and repeated them across three channels, creating a 3-channel 2D image in an RGB-like format. To balance the dataset, we just augmented the AD group by applying random rotations within a range of -5° to +5°.
\subsection{Experimental Setup}
In all experiments, we initially used the holdout method for data splitting with an 8:2 ratio. This approach was chosen due to limited computational resources and the considerable time required to train the models. Once the holdout experiments were completed, we applied 5-fold cross-validation using the best settings identified from the holdout phase. To evaluate the effectiveness of ViTLoRA, we conducted experiments with LoRA matrices placed in different configurations:
\begin{enumerate}
	\item All Encoders: Adaptation matrices were added to the key, query, and value matrices in all encoder blocks.
	\item Last Two Encoders: Adaptation matrices were added only to the last two encoder blocks for adaptation to key, query, and value matrices.
\end{enumerate}

We evaluated LoRA matrices with varying ranks in each configuration. Table \ref{tab:rankcomparison} presents the results of these experiments, including the training and validation accuracies for each setup. Our findings indicate that the best performance was achieved using adaptation matrices with a rank of 4 for training the ViTLoRA model, and a rank of 8 for the EffNetViTLoRA model. In both cases, the low-rank matrices were applied in all encoder blocks, adapting the key, query, and value weight matrices. This configuration resulted in a highly efficient model with only 419K learnable parameters in EffNetViTLoRA, a significant reduction compared to training the full ViT model from scratch. These results highlight the efficiency and adaptability of the ViTLoRA approach, making it a promising solution for applying Vision Transformers to medical imaging tasks, where data availability is often limited. We also evaluated feature maps from various blocks of the pretrained EfficientNet to connect to the pretrained ViTLoRA in our hybrid model. The best performance was achieved using the output from the final block, which has a spatial resolution of 12 × 12.

Our experiments were conducted on an NVIDIA RTX 3090 GPU \cite{cuda}. The number of training epochs was set to 100, determined through iterative experimentation and performance monitoring. The batch size was fixed at 32. To optimize the models, we used the Adam optimizer \cite{kingma2014adam}, with a learning rate of 1e-3 for the ViTLoRA model and 1e-4 for both EfficientNet and EffNetViTLoRA.
\begin{table}
	\scriptsize
	\caption{Comparison of using different ranks in configurations 1 and 2 using hold-out split method with 8:2 ratio. The best result obtained by Configuration 1 with rank of 4. All hyper parameters were same in all experiments.}
	\centering
	\label{tab:rankcomparison}
	\setlength{\tabcolsep}{3pt}
	\begin{tabular}{lcccc} 
		\toprule
		\textbf{Rank} & \multicolumn{2}{c}{\textbf{Configuration 1}} & \multicolumn{2}{c}{\textbf{Configuration 2}} \\
		\cmidrule(lr){2-3} \cmidrule(lr){4-5}
		& train accuracy & validation accuracy & train accuracy & validation accuracy \\
		\midrule
		R = 1        & 95\%      & 82\%      & 98\%      & 82\%      \\
		R = 2        & 98\%      & 85\%      & 97\%      & 80\%      \\
		R = 4        & 96\%     & \textbf{86\%}      & 96\%      & 80\%      \\
		R = 16        & 90\%      & 75\%      & 97\%      & 83\%      \\
		R = 32        & 62\%      & 54\%     & 87\%      & 75\%     \\
		\bottomrule
	\end{tabular}
\end{table}	
\subsection{Evaluation Metrics}
We evaluated the model's performance using common metrics, including accuracy, precision, recall, and F1-score, which are mathematically defined as follows:
\begin{equation}
	Accuracy = \frac {TP + TN}{TP + TN + FP + FN}
\end{equation}
\begin{equation}
	Precision = \frac{TP}{TP + FP}
\end{equation}
\begin{equation}
	Recall = \frac{TP}{TP + FN}
\end{equation}
\begin{equation}
	\text{F1-score} = 2 \times \frac{Precision \times Recall}{Precision + Recall}
\end{equation}
Where TP, TN, FP, and FN stand for True Positive, True Negative, False Positive, and False Negative, respectively.
In medical research, recall is particularly important, as a high recall indicates that a larger proportion of true positive cases are correctly identified, an essential factor in the diagnosis of Alzheimer’s disease.
\subsection{Results}
\begin{figure*}[!t]
	\centering
	\begin{subfigure}{0.45\textwidth}
		\centering
		\includegraphics[width=\textwidth]{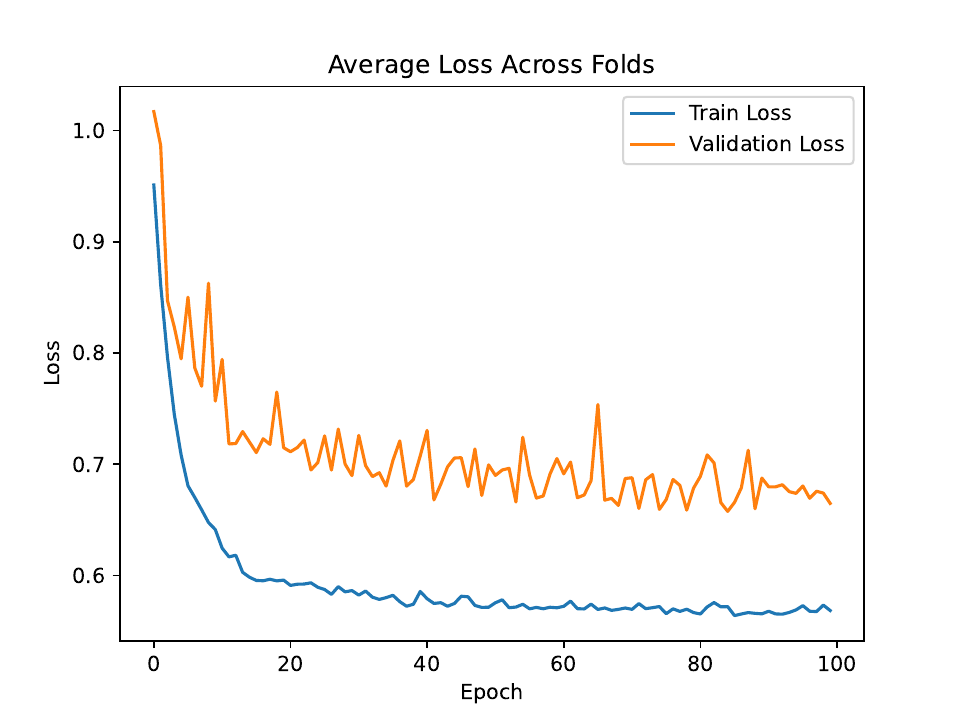}
		\caption{}
	\end{subfigure}
	\begin{subfigure}{0.45\textwidth}
		\centering
		\includegraphics[width=\textwidth]{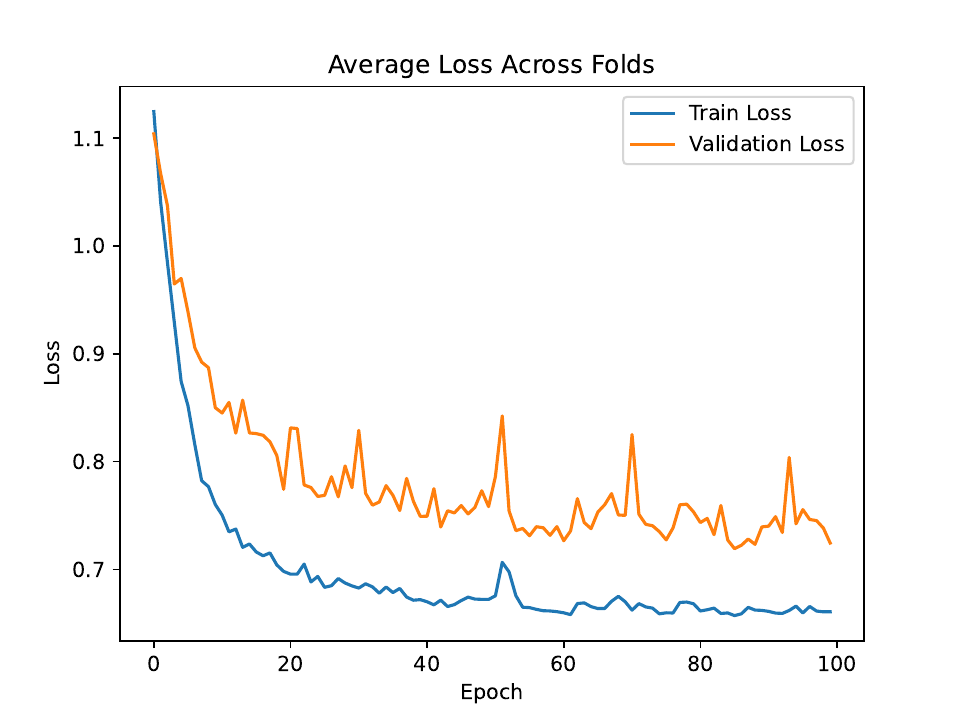}
		\caption{}
	\end{subfigure}
	\begin{subfigure}{0.45\textwidth}
		\centering
		\includegraphics[width=\textwidth]{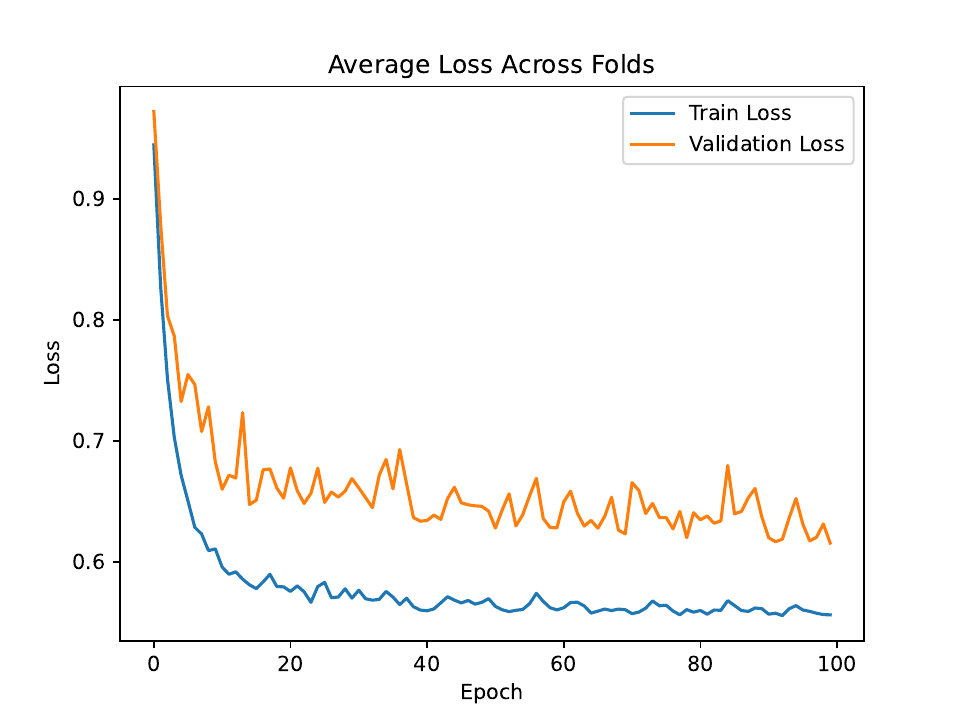}
		\caption{}
	\end{subfigure}
	\caption{Plots of train and validation losses using 5-fold cross-validation. (a) ViTLoRA. (b) EfficientNet. (c) EffNetViTLoRA.}
	\label{fig:loss_plots}
\end{figure*}

\begin{figure*}[!t]
	\centering
	\begin{subfigure}{0.45\textwidth}
		\centering
		\includegraphics[width=\textwidth]{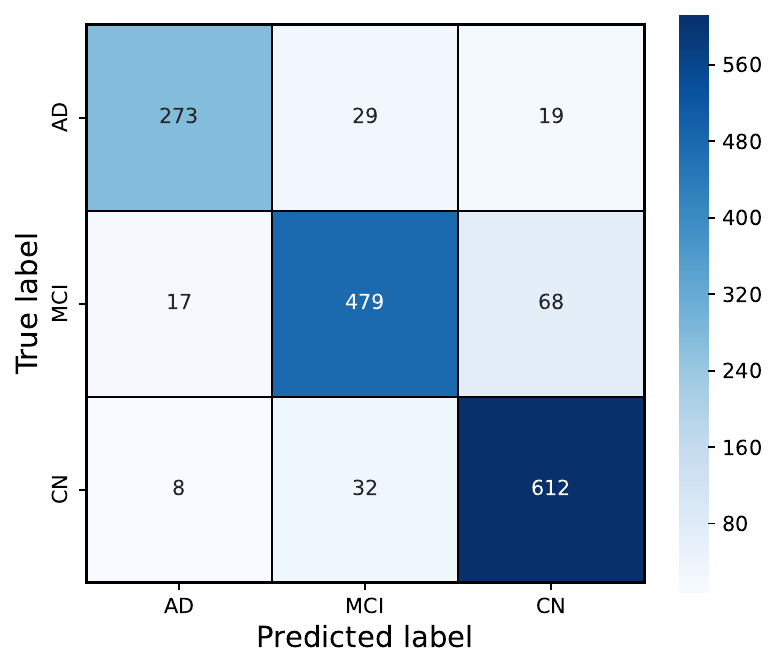}
		\caption{}
	\end{subfigure}
	\begin{subfigure}{0.45\textwidth}
		\centering
		\includegraphics[width=\textwidth]{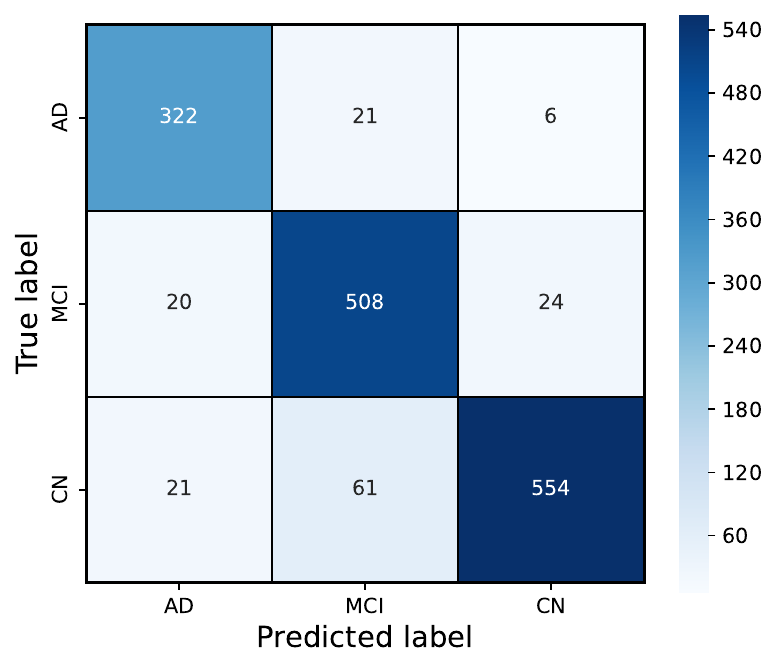}
		\caption{}
	\end{subfigure}
	\begin{subfigure}{0.45\textwidth}
		\centering
		\includegraphics[width=\textwidth]{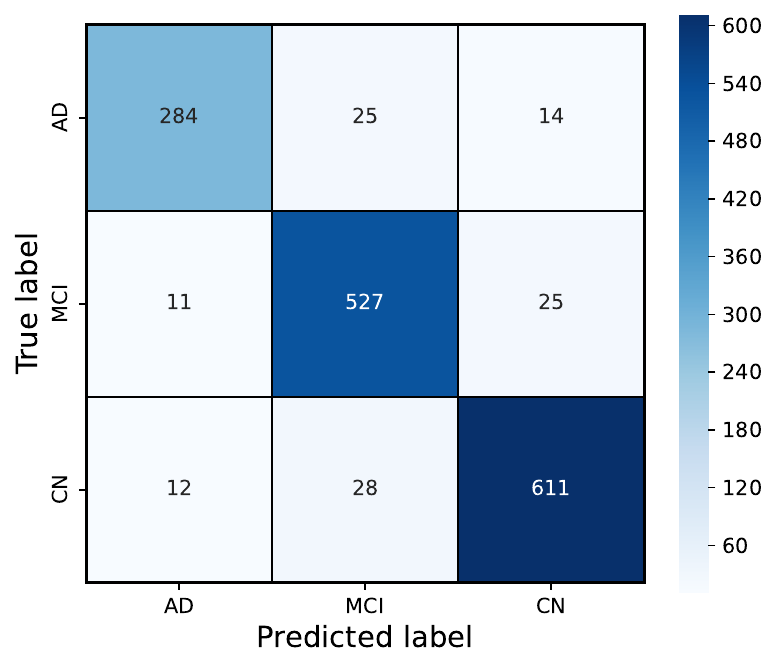}
		\caption{}
	\end{subfigure}
	\caption{Confusion matrices for the best-performing models in 5-fold cross-validation. (a) ViTLoRA. (b) EfficientNet. (c) EffNetViTLoRA. Although MCI falls between the other two groups (CN and AD), and is difficult to diagnose, the hybrid Model (c) outperformed the other two models in accurately identifying the MCI cases.}
	\label{fig:conf_matrix_plots}
\end{figure*}

\begin{figure*}[!t]
	\centering
	\begin{subfigure}{0.45\textwidth}
		\centering
		\includegraphics[width=\textwidth]{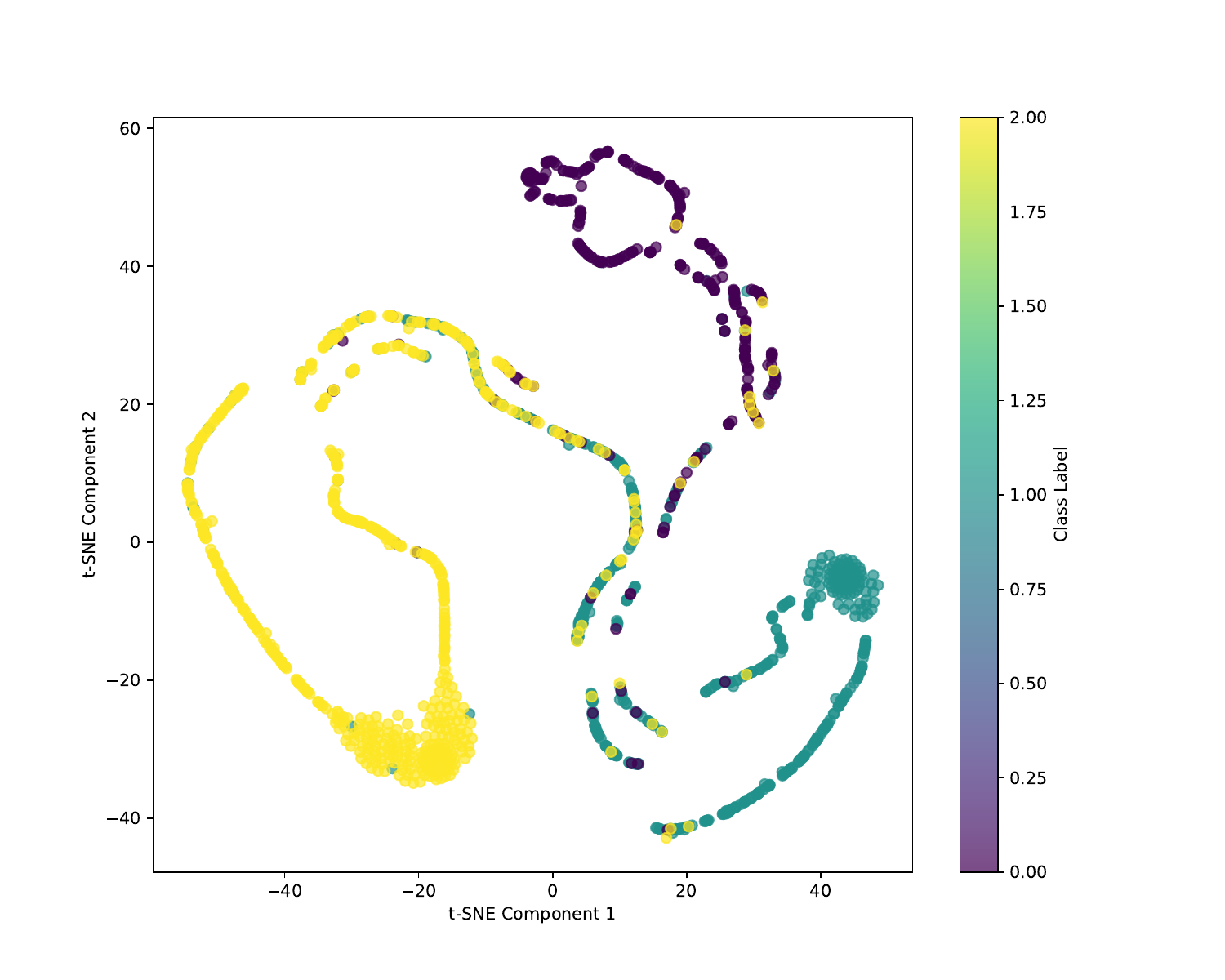}
		\caption{}
	\end{subfigure}
	\begin{subfigure}{0.45\textwidth}
		\centering
		\includegraphics[width=\textwidth]{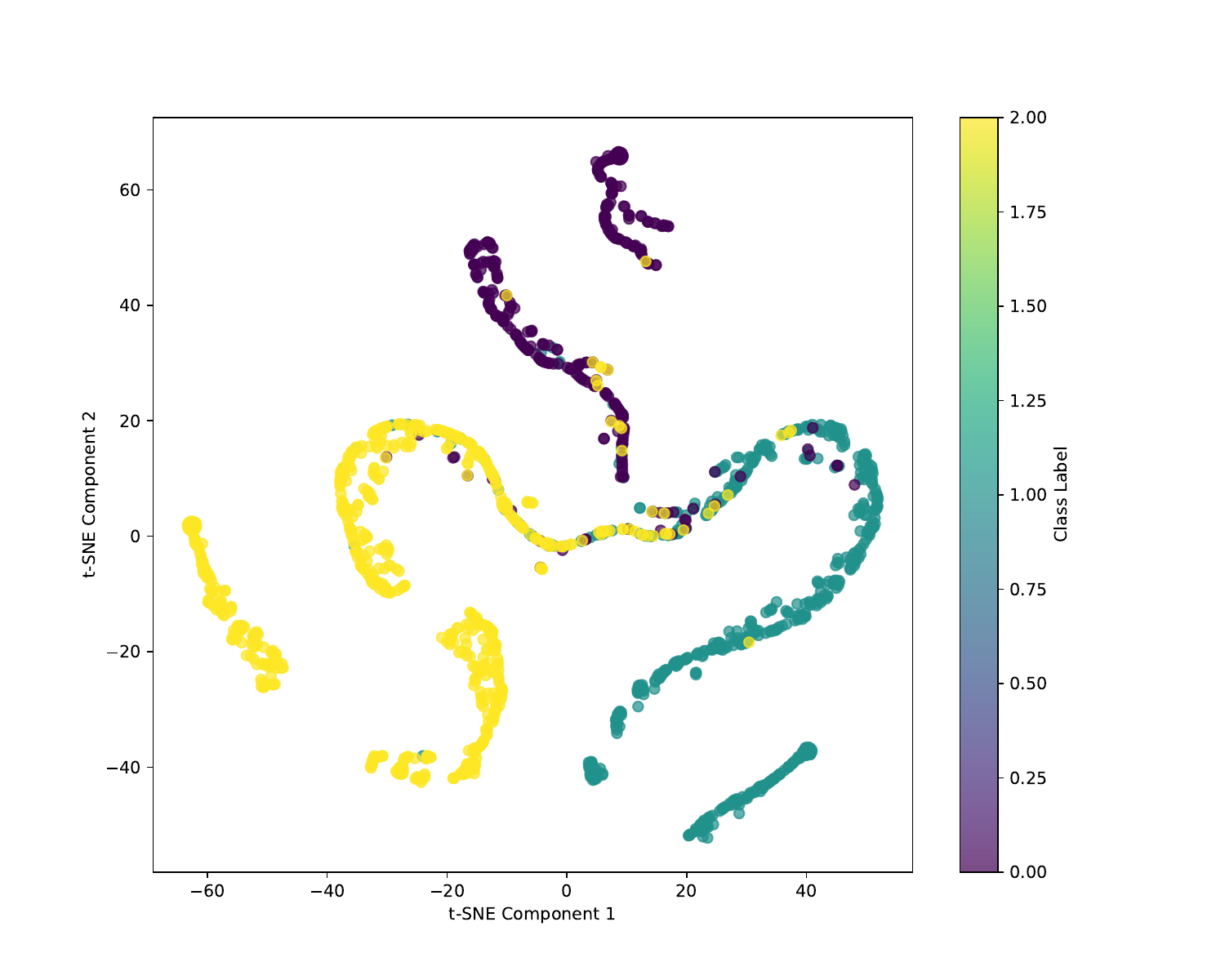}
		\caption{}
	\end{subfigure}
	\begin{subfigure}{0.45\textwidth}
		\centering
		\includegraphics[width=\textwidth]{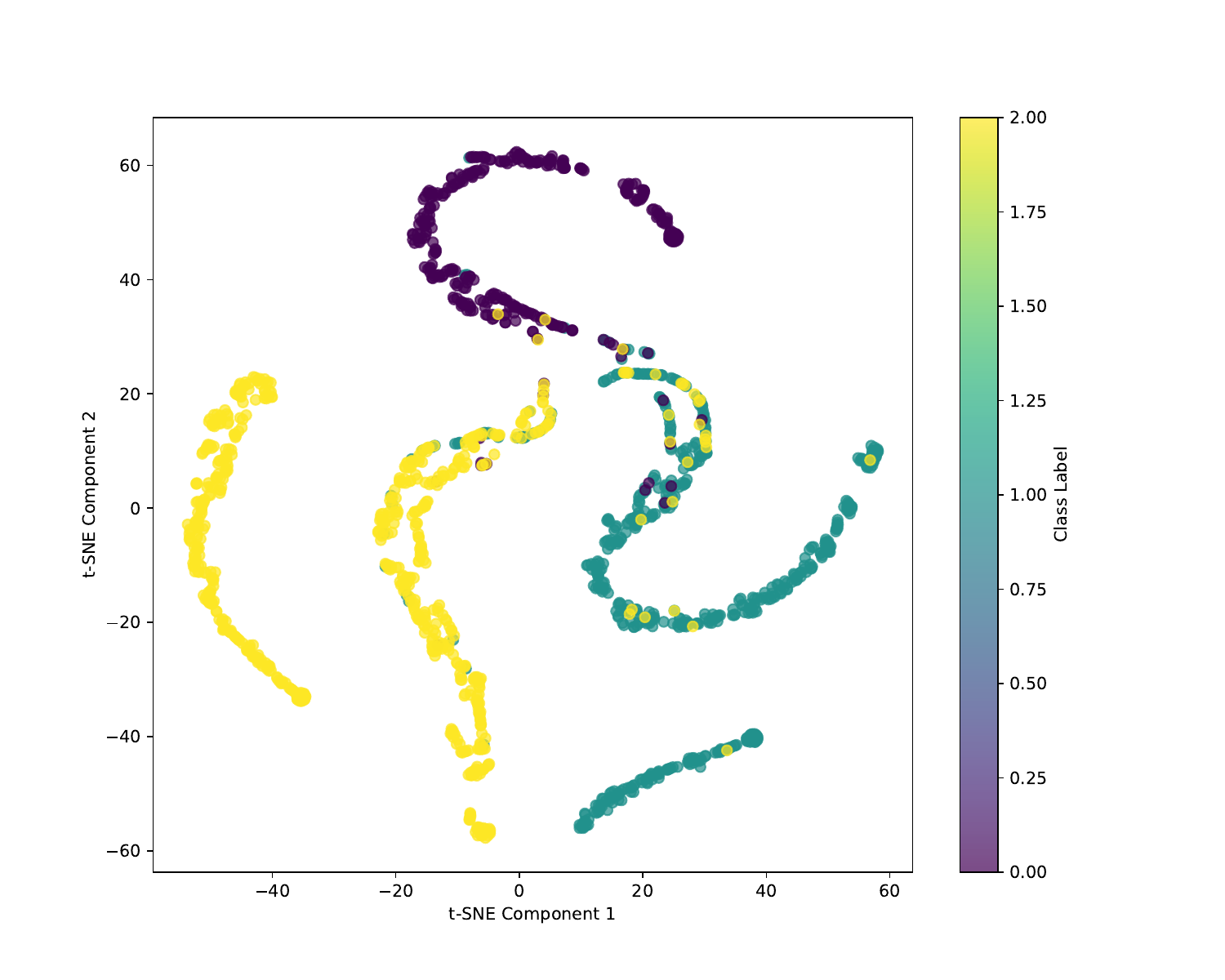}
		\caption{}
	\end{subfigure}
	\caption{Comparison of t-SNE feature visualization using two components. (a) ViTLoRA. (b) EfficientNetV2. (c) EffNetViTLoRA. The hybrid model (c) performs better in feature extraction and separating samples with least coincidence than other two models.}
	\label{fig:tSNE_plots}
\end{figure*}

Table \ref{tab:metrics_results} summarizes the average evaluation metrics. The hybrid model, EffNetViTLoRA, outperforms both the ViTLoRA and EfficientNetV2 models across all metrics. This demonstrates that combining local and global features representing local and global dependencies within an Image, leads to superior performance in AD diagnosis. We also put results of the other works which have done three class classification, to make a comparison with our work. Note that other studies may achieve better classification results, but they have just used a subset of MRI volumes which means there is a bias selection in their datasets. Fig.~\ref{fig:loss_plots} shows the average training and validation loss plots for ViTLoRA, EfficientNetV2, and the Hybrid model, obtained through 5-fold cross-validation. We did not continue training more than 100 epochs, because there was not any improvement in loss reduction. Additionally, we plotted confusion matrices of the best-performing models to determine which class has been misclassified with other classes. As Fig.~\ref{fig:conf_matrix_plots} demostrates, in all three models, MCI class discrimination is challenging as it is a midlevel class between CN and AD. The MCI subjects may be misdiagnosed as CN or AD groups. To further illustrate the hybrid model's ability to discriminate between classes, we applied t-SNE visualization to the extracted features. Fig.~\ref{fig:tSNE_plots} presents the 2D t-SNE plot, which highlights the model’s strong ability to distinguish feature classes across three groups. It is worth noting that distinguishing MCI cases remains challenging in some instances, as their features may overlap with those of CN or AD groups. This is not unexpected, given that our dataset includes ALL MRI subjects from ADNI without bias selection or any assumption in selecting data items. Furthermore, MCI inherently represents a transitional state between Cognitively Normal and Alzheimer’s Disease states, making the classification more complex.
\begin{table*}
	\caption{Comparison of metrics across proposed methods and other studies for three class classification of Alzheimer's disease, using 5-fold cross-validation. Unlike previous studies, we used ALL MRI T1-weighted data available by the time of our research.}
	\centering
	\label{tab:metrics_results}
	\renewcommand{\arraystretch}{1.2}
	\setlength{\tabcolsep}{4pt} 
	\resizebox{\textwidth}{!}{ 
	\begin{tabular}{@{}lp{2.5cm}p{2cm}cccccc@{}}
		\toprule
		\textbf{Method}      & \textbf{Dataset} & \textbf{\# of Subjects} & \textbf{Multimodality} & \textbf{Accuracy (\%)} & \textbf{Precision (\%)} & \textbf{Recall (\%)} & \textbf{F1 Score (\%)} \\ \midrule
		ViTLoRA (rank = 4)             & ADNI    & \textbf{2010}                   & $\times$             & 88.74                  & 88.82                   & 88.74                & 88.78                  \\
		EfficientNetV2 small        & ADNI    & \textbf{2010}                 & $\times$                 & 90.05                  & 90.10                   & 90.08                & 90.10                  \\
		EffNetViTLoRA (hybrid model) & ADNI    & \textbf{2010}                   & $\times$             & {92.52}         & {92.66}          & {92.87}       & {92.76}         \\
		\cite{lin2021multiclass}                      & ADNI    & 746                   & $\checkmark$                 & 66.7                   & -                    & -                 & 64.9                   \\
		\cite{altaf2018multi}                      & ADNI    & 287                   & $\checkmark$                 & 79.8                   & 79.0                    & 92.0                & -                   \\
		\cite{khan2019transfer}                      & ADNI    & 150                   & $\times$                 & 95.19                   & -                    & -                & -                   \\
		\cite{raju2020multi}                      & ADNI    & 465                   & $\times$                 & 97.77                   & 97.93                    & 97.76                & 97.80                   \\
		\cite{raju2021multi}                      & ADNI    & 465                   & $\times$                 & 96.66                   & 96.70                    & 96.66                & 96.66                   \\
		\cite{jain2019convolutional}                      & ADNI    & 150                   & $\times$                 & 95.73                   & 96.33                    & 96.0                &   95.66                 \\
		\cite{olatunde2024multiclass}                      & ADNI    & 194                   & $\checkmark$                 & 84.87                  & -                    & -               &   84.83                 \\
		\cite{golovanevsky2022multimodal}                      & ADNI    & 239                   & $\checkmark$                 & 96.88                  & -                    & -               &   91.41                 \\
		\bottomrule
	\end{tabular}}
\end{table*} 
\section{Conclusion}
\label{sec:conclusion}
In this paper, we proposed EffNetViTLoRA, a hybrid model for diagnosing Alzheimer’s disease across three classes: AD, MCI, and CN. The model combines EfficientNetV2 for extracting local spatial features with ViTLoRA, an adapted Vision Transformer pretrained on ImageNet-1K that employs low-rank decomposable matrices to capture differences between source and target datasets without altering pretrained weights.
The integration of local and global feature extraction significantly improved classification performance, demonstrating the potential of hybrid architectures in medical imaging. To ensure generalizability, we used MRI data from ALL subjects in the baseline visit of the ADNI dataset, including ADNI1, ADNIGO, ADNI2, and ADNI3 phases, without any bias selection. EffNetViTLoRA achieved over 92\% in accuracy, precision, and recall, demonstrating its potential as a novel and reliable diagnostic model. The code of this project is publicly available at [Anonymous Link (journal constraint)].

Future work will explore incorporating specific brain regions such as segmented gray matter, white matter, and hippocampi, known to be affected in Alzheimer’s disease, to further enhance model performance.

  \bibliographystyle{elsarticle-num} 
  \bibliography{library}

\end{document}